# Multi-output Classification Framework and Frequency Layer Normalization for Compound Fault Diagnosis in Motor

Wonjun Yi*[†], Yong-Hwa Park**
*Dept. of Electrical Eng., KAIST., ** Dept. of Mechanical Eng., KAIST.

**Abstract**

This work introduces a multi-output classification (MOC) framework designed for domain adaptation in fault diagnosis, particularly under partially labeled (PL) target domain scenarios and compound fault conditions in rotating machinery. Unlike traditional multi-class classification (MCC) methods that treat each fault combination as a distinct class, the proposed approach independently estimates the severity of each fault type, improving both interpretability and diagnostic accuracy. The model incorporates multi-kernel maximum mean discrepancy (MK-MMD) and entropy minimization (EM) losses to facilitate feature transfer from the source to the target domain. In addition, frequency layer normalization (FLN) is applied to preserve structural properties in the frequency domain, which are strongly influenced by system dynamics and are often stationary with respect to changes in rpm. Evaluations across six domain adaptation cases with PL data demonstrate that MOC outperforms baseline models in macro F1 score. Moreover, MOC consistently achieves better classification performance for individual fault types, and FLN shows superior adaptability compared to other normalization techniques.

## 1. Introduction

Compound fault diagnosis in rotating machinery plays a critical role in ensuring system reliability. However, the presence of domain shifts and scarcity of labeled data often limits the performance and adaptability of data-driven models, especially under unsupervised domain adaptation (UDA) settings. Traditional multi-class classification (MCC) approaches tend to fall short in such environments, as they rely on a single-label representation that struggles to capture the complex nature of coexisting faults and cannot effectively handle partially labeled (PL) target domains. Some researchers tried to develop MCC classifier for compound fault severity level classification [1,2] but, since their problem dealt with only seven [1] to thirteen classes [2], there were less needs for developing alternative architecture for compound fault diagnosis. Other researchers tried to develop multi-label classification (MLC) approaches for compound fault diagnosis [3,4], but in this case, we can only determine whether each fault is presence or absence, but can not classify exact severity level.

To overcome these limitations, in this work, we build upon our previously published study [5], which first introduced the multi-output classification (MOC) framework for fault diagnosis under PL target conditions. The current paper presents an extended and more comprehensive investigation, offering refined analyses and enhanced explanations for key components of the original framework. Specifically, the proposed MOC model assigns each fault type to a distinct task-specific layer (TSL), allowing it to isolate classification tasks, mitigate inter-class interference, and improve domain-invariant feature representation. The model structure is inspired by the tasks-constrained deep convolutional network (TCDCN) [6], which has demonstrated efficacy in multi-task learning scenarios through the use of independent TSLs.

[†] Wonjun Yi, lasscap@kaist.ac.kr

To improve robustness under varying operating conditions, we introduce frequency layer normalization (FLN), a normalization method tailored for vibration signals in the frequency domain. FLN preserves domain-stable spectral characteristics and is particularly effective in countering distributional shifts induced by rpm variations and torque fluctuations. Unlike conventional batch normalization (BN) [7], layer normalization (LN) [8], and instance normalization (IN) [9], which normalize over batch statistics or general feature dimensions, FLN applies normalization strictly along the frequency axis, capturing frequency-dominant patterns more reliably.

Experiments were conducted using a benchmark motor dataset that includes four types of faults: inner race fault (IRF), outer race fault (ORF), misalignment, and unbalance. Data were collected under three different operational profiles defined by rpm patterns and torque loads. In total, six domain adaptation configurations were evaluated using the macro F1 score to assess classification performance. Through this extended study, we provide a deeper evaluation of the MOC framework and demonstrate how FLN contributes to improved diagnostic performance under partially labeled domain settings.

## 2. Theory and Experiment

### 2.1 Multi-Output Classification

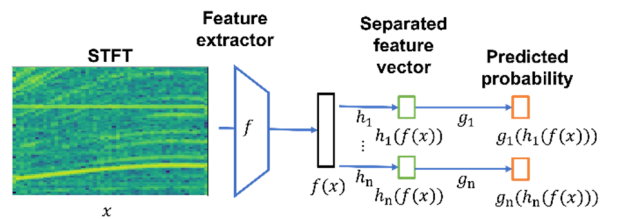

**Fig. 1** Architecture of multi-output classification.

Prior research [10] has demonstrated that jointly learning classification tasks that share underlying physical relationships can lead to improved overall performance. This is attributed to the fact that such related tasks—such as, for example, the maneuvering direction and environmental conditions of a drone—encourage the feature extractor to learn more physically grounded representations, as opposed to overfitting to task-specific artifacts.

Given that the constituent faults in a compound fault scenario often influence each other dynamically [11, 12], the multi-output classification (MOC) framework is also well-suited to this context. We propose a MOC architecture derived from the tasks-constrained deep convolutional network (TCDCN) [6], as depicted in Fig. 1. In a conventional multi-class classification (MCC) framework, a specific combination of fault severity levels—such as inner race fault (IRF) of 0.2 mm, outer race fault (ORF) of 0.2 mm, misalignment of 0.15 mm, and unbalance of 10.034 g—is treated as a single, unique class. In contrast, MOC assigns each fault type to a separate task-specific layer (TSL), enabling the model to evaluate severity independently for each fault dimension.

While MCC encodes all fault information jointly, it maps each compound condition to an atomic class without reflecting the physical relationships among fault types. This often leads to feature entanglement and poor generalization. In contrast, MOC uses task-specific layers to separately classify each fault type, making the model better aligned with the physical structure of compound faults and more robust under domain shifts.

Beyond the benefits established in [10], the MOC framework provides additional advantages over MCC in compound fault settings. As the number of discrete severity levels for each fault type increases, the total number of possible fault combinations grows multiplicatively. This results in an exponential expansion of the class space in MCC, which becomes increasingly difficult to model effectively. Moreover, under UDA conditions, using a shared feature extractor to align features across such a complex and densely packed class distribution is inherently problematic and often infeasible.

In this configuration, the component denoted as $h$ in Fig. 1 is responsible for performing feature alignment through unsupervised domain adaptation (UDA) loss functions, such as multi-kernel maximum mean discrepancy (MK-MMD) [13], while the $g$ component serves as a SoftMax-based classifier, trained using categorical cross-entropy (CCE) along with entropy minimization (EM) loss [14].

## 2.2 Frequency Layer Normalization

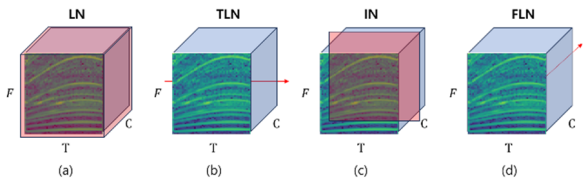

**Fig. 2** Conventional and proposed normalization methods: (a) LN, (b) TLN, (c) IN, (d) FLN (proposed).

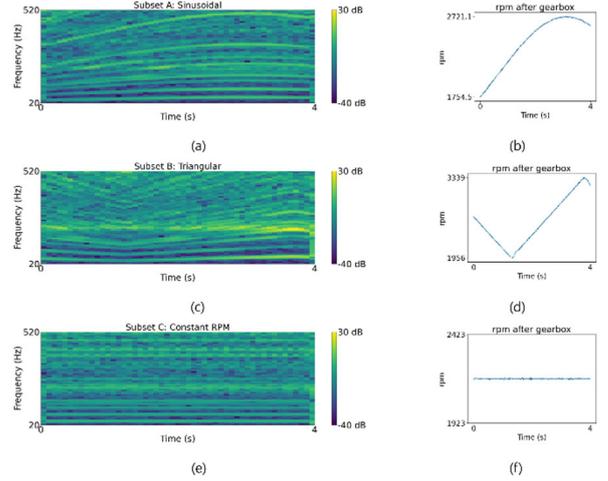

**Fig. 3** STFT and rpm after gearbox. (a), (b) under sinusoidal rpm pattern, (c), (d) under triangular rpm pattern, (e), (f) under constant rpm.

We propose frequency layer normalization (FLN), a normalization strategy specifically designed to align with the structural and spectral properties of vibration signals generated by rotating machinery. These signals typically exhibit dominant frequency components resulting from periodic mechanical excitations. Common fault types—including imbalance, misalignment, and bearing defects such as inner race fault (IRF) and outer race fault (ORF)—give rise to distinguishable spectral signatures that correlate with fundamental machine dynamics.

Each type of fault produces frequency components that can be directly mapped to specific physical causes. For example, imbalance typically generates a strong frequency component at 1× shaft speed, while misalignment often produces significant energy at 2× shaft speed and its harmonics. IRF and ORF, in contrast, are known to excite bearing-specific resonance frequencies such as ball pass frequency inner (BPFI), $f_{BPFI}$, and ball pass frequency outer (BPFO), $f_{BPFO}$, which are defined as:

$$f_{BPFI} = \frac{n}{2}(1 + \frac{d}{D}\cos\theta)f_r \qquad (1)$$

$$f_{BPFO} = \frac{n}{2}(1 - \frac{d}{D}\cos\theta)f_r \qquad (2)$$

where $n$ is the number of rolling elements, $d$ and $D$ are the ball diameter and pitch diameter, respectively, $\theta$ is the contact angle, and $f_r$ is the shaft rotation frequency (rpm-based). These fault-related components vary linearly with rpm and thus result in frequency-domain patterns that shift predictably as operating speed changes. This linear dependence renders the frequency axis not only physically meaningful but also stable under varying conditions, making it an ideal basis for normalization in domain adaptation.

FLN capitalizes on this characteristic by computing normalization statistics exclusively along the frequency axis. This axis-aligned approach allows the model to preserve the spectral features associated with mechanical faults. In contrast, conventional layer normalization (LN) [8] computes statistics across all features in a sample indiscriminately, which can disrupt physically relevant patterns. To test the hypothesis that frequency-aware normalization yields superior domain transfer performance, we conducted an ablation study using time layer normalization (TLN), where normalization is instead applied along the time axis. This configuration neglects the spectral regularities induced by machine dynamics and produced inferior results in our experiments.

Additional comparisons were made with batch normalization (BN) [7] and instance normalization (IN) [9]. While BN computes statistics across batches and is widely used in general-purpose deep learning, it is known to be sensitive to batch composition and less effective in domain adaptation settings. IN, on the other hand, performs per-sample, per-channel normalization and has demonstrated effectiveness in style transfer tasks by isolating structural content from stylistic variation. However, its lack of alignment with the deterministic frequency signatures in mechanical systems limits its utility in fault diagnosis scenarios.

To illustrate these differences, Fig. 2 presents a schematic comparison of FLN and other normalization techniques, including LN, TLN, and IN. Each blue cube in the figure represents an input sample, and red arrows or shaded areas highlight the area which statistical normalization is performed. For example, In Fig. 2 (d), FLN is shown to normalize along the frequency axis ($F$). In other words, for same batch, channel, and time index, frequency values are normalized with same mean and variance.

To further support the physical validity of this approach, Fig. 3 visualizes the short-time Fourier transform (STFT) of vibration signals across varying rpm conditions. This shows STFT of motor system is frequency dominant. Therefore, this observation underscores the effectiveness of FLN in stabilizing feature distributions under domain shifts induced by operational condition differences.

### 2.3 Experimental Setup

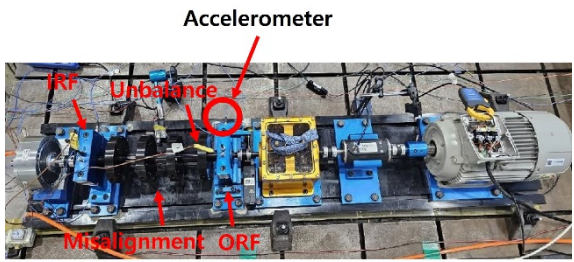

**Fig. 4** Motor used for the experiment

During domain adaptation, the model was pre-trained

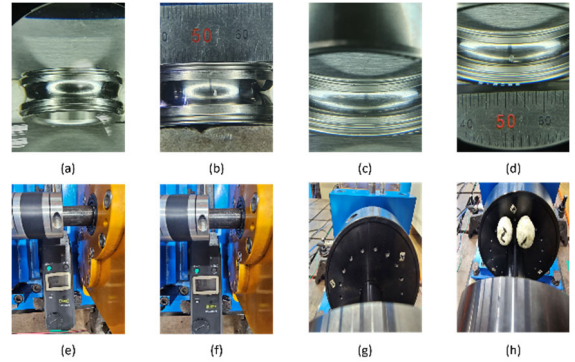

**Fig. 5** Induced faults. (a) Normal inner race, (b) IRF 0.2 mm, (c) Normal outer race, (d) ORF 0.2 mm, (e) Normal (No misalignment), (f) Misalignment 0.15 mm, (g) Normal (No unbalance), (h) Unbalance 18.070 g

Experiments were conducted using a custom-designed rotating machinery testbed [15] configured to simulate compound fault conditions and domain adaptation scenarios. As shown in Fig. 4, the system includes a three-phase induction motor, gearbox, two bearing housings, a shaft, rotor components, a torque meter, and a hysteresis brake. A single piezoelectric accelerometer (PCB352C34) was installed on one of the bearing housings in the x-direction, following the guidelines of ISO 10816-1:1995 [16]. The sensor location is indicated by the red circle in Fig. 4. The testbed utilized standardized deep-groove ball bearings (NSK 6205 DDU). Key bearing specifications are: ball diameter $d = 7.90\ mm$, pitch diameter, $D = 38.5\ mm$, contact angle $\theta = 0$, and number of rolling elements $N = 9$. Mechanical loading was applied using a hysteresis brake (AHB-3A, Valid Magnetic Ltd.). Notably, the hysteresis brake was only used in Subset A to simulate dynamic loading conditions. Torque was monitored using a torque meter (M425, Datum Electronics).To induce compound faults, inner race fault (IRF) and outer race fault (ORF) were introduced into the bearing components, while misalignment and unbalance were applied through modifications to the rotor assembly. Severity levels were set at 0 mm and 0.2 mm for IRF and ORF, 0 mm, 0.15 mm, and 0.3 mm for misalignment, and 0 g, 10.034 g, and 18.070 g for unbalance. The physical manifestations of these faults are illustrated in Fig. 5. These settings resulted in a total of 36 distinct compound fault conditions.

To simulate domain adaptation scenarios, three operational subsets were defined: Subset A featured a sinusoidal rpm profile and manually randomized torque loads, Subset B followed a triangular rpm pattern, and Subset C maintained a constant rpm with no applied load. For each experimental trial, vibration data were sampled at 25.6 kHz over a 4-second window. Fault condition data represented only 10% of the quantity of normal condition data. During domain adaptation experiments, the source domain dataset consisted entirely of labeled samples, while the target domain dataset contained a mix of 10% labeled and 90% unlabeled data, representing a partially labeled (PL) setting.

During domain adaptation, the model was first pre-trained on the source domain for 100 epochs and subsequently fine-tuned on the target domain for an additional 100 epochs. Optimization was performed using the Adam [17] with a fixed learning rate of 1e-3. At each epoch, the model parameters corresponding to the lowest validation loss were selected as the final model.

2.4 Results

**Table 1** Macro F1 score based on model architecture

| Source → Target | MCC | MOC |
|---|---|---|
| A → B | 0.991 | 0.999 |
| A → C | 0.673 | 0.774 |
| B → A | 0.827 | 0.807 |
| B → C | 0.833 | 0.870 |
| C → A | 0.766 | 0.797 |
| C → B | 1.000 | 0.996 |
| Average | 0.848 | 0.874 |

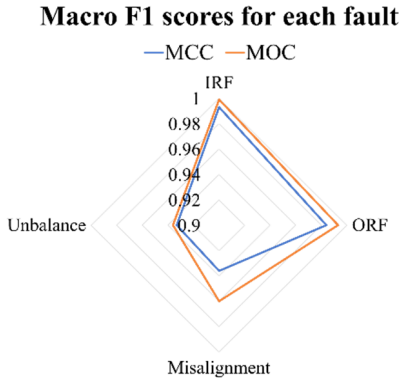

**Fig. 6** Macro F1 scores for each fault
(averaged for all domain adaptation scenarios)

As presented in Table 1, the multi-output classification (MOC) framework consistently yields higher macro F1 scores than the conventional multi-class classification (MCC) approach across all domain adaptation cases. In addition, Fig. 4 illustrates that MOC provides improved classification performance for each individual fault category. This performance gain is attributed to the use of task-specific layers (TSLs), which enable the model to independently estimate the severity of each fault. This modular structure facilitates more accurate fault-level discrimination and leads to more effective diagnosis of compound fault scenarios.

Moreover, as shown in Table 2, FLN outperforms other conventional normalization methods such as BN [7], LN [8], TLN, and IN [9], when integrated within the same multi-output classification (MOC) architecture, highlighting its effectiveness in enhancing domain adaptation performance.

**Table 2** Macro F1 score based on normalization method

| Source → Target | BN [7] | LN [8] | TLN [8] | IN [9] | FLN |
|---|---|---|---|---|---|
| A → B | 0.984 | 0.805 | 0.691 | 0.959 | 0.999 |
| A → C | 0.810 | 0.623 | 0.093 | 0.801 | 0.774 |
| B → A | 0.661 | 0.588 | 0.439 | 0.715 | 0.807 |
| B → C | 0.782 | 0.699 | 0.077 | 0.820 | 0.870 |
| C → A | 0.705 | 0.574 | 0.263 | 0.737 | 0.797 |
| C → B | 0.960 | 0.864 | 0.591 | 0.945 | 0.996 |
| Average | 0.817 | 0.692 | 0.359 | 0.829 | 0.874 |

## 3. Conclusion

The proposed MOC framework achieved strong performance in domain adaptation tasks, effectively addressing challenges associated with compound faults and PL data. Additionally, FLN consistently outperformed BN, LN, TLN, and IN by exploiting frequency-domain characteristics, leading to improved adaptability under varying operational conditions.

Future work may explore more advanced architectural designs to further enhance the robustness and scalability of MOC.

## References

(1) Amanda Rosa Ferreira Jorge et al. "Rotodynamics Multi-Fault Diagnosis through Time Domain Parameter Analysis with MLP: A Comprehensive Study". In: 2024 International Workshop on Artificial Intelligence and Machine Learning for Energy Transformation (AIE). IEEE. 2024, pp. 1–6.
(2) Ruben Medina et al. "Scale-Fractal Detrended Fluctuation Analysis for Fault Diagnosis of a Centrifugal Pump and a Reciprocating Compressor". In: Sensors 24.2 (2024).
(3) Yanrui Jin et al. "Actual bearing compound fault diagnosis based on active learning and decoupling attentional residual network". In: Measurement 173 (2021), p. 108500.
(4) Liuxing Chu et al. "Exploring the essence of compound fault diagnosis: A novel multi-label domain adaptation method and its application to bearings". In: Heliyon 9.3 (2023).
(5) Yi, W., and Park, Y.-H., 2025, "Multi-output Classification for Compound Fault Diagnosis in Motor under Partially Labeled Target Domain,"
(6) Zhang, Z., Luo, P., Loy, C. C., and Tang, X., 2014, "Facial Landmark Detection by Deep Multi-Task Learning," Computer Vision--ECCV 2014: 13th European Conference, Zurich, Switzerland, September 6-12, 2014, Proceedings, Part VI 13, Springer, pp. 94-108.
(7) Ioffe, S., and Szegedy, C., 2015, "Batch Normalization: Accelerating Deep Network Training by Reducing Internal Covariate Shift," International Conference on Machine Learning, PMLR, pp. 448-456.
(8) Ba, J. L., Kiros, J. R., and Hinton, G. E., 2016, "Layer Normalization,"
(9) Ulyanov, D., Vedaldi, A., and Lempitsky, V., 2016, "Instance Normalization: The Missing Ingredient for Fast Stylization,"
(10) Yi, W., Choi, J.-W., and Lee, J.-W., 2023, "Sound-Based Drone Fault Classification Using Multi-Task Learning," *Proceedings of the 29th International Congress on Sound and Vibration (ICSV29)*, International Institute of Acoustics and Vibration.
(11) Pandya, D. H., Upadhyay, S. H., and Harsha, S. P., 2014, "Nonlinear Dynamic Analysis of High Speed Bearings due to Combined Localized Defects,"

*Journal of Vibration and Control*, Vol. 20, No. 15, pp. 2300–2313.

(12) Randall, R. B., and Antoni, J., 2011, "Rolling Element Bearing Diagnostics—A Tutorial," *Mechanical Systems and Signal Processing*, Vol. 25, No. 2, pp. 485–520.

(13) Long, M., Cao, Y., Wang, J., and Jordan, M., 2015, "Learning Transferable Features with Deep Adaptation Networks," *International Conference on Machine Learning*, PMLR, pp. 97-105.

(14) Grandvalet, Y., and Bengio, Y., 2004, "Semi-Supervised Learning by Entropy Minimization," *Advances in Neural Information Processing Systems*, Vol. 17.

(15) Jung, W., Kim, S. H., Yun, S. H., Bae, J., and Park, Y. H., 2023, "Vibration, Acoustic, Temperature, and Motor Current Dataset of Rotating Machine under Varying Operating Conditions for Fault Diagnosis," Data in Brief, Vol. 48, Elsevier, pp. 109049.

(16) International Organization for Standardization. ISO 10816-1:1995 Mechanical vibration – Evaluation of machine vibration by measurements on non-rotating parts. Tech. rep. Geneva, Switzerland: International Organization for Standardization (ISO), 1995.

(17) Kingma, D. P., and Ba, J., 2014, "Adam: A Method for Stochastic Optimization,"